\pdfoutput=1

\documentclass[11pt]{article}

\usepackage[final]{acl}

\usepackage{times}
\usepackage{latexsym}

\usepackage{listings}
\usepackage{lipsum} 

\usepackage[T1]{fontenc}

\usepackage[utf8]{inputenc}

\usepackage{microtype}

\usepackage{inconsolata}

\usepackage{graphicx}

\usepackage{microtype}
\usepackage{hyperref}
\usepackage{url}
\usepackage{booktabs}

\usepackage{multicol}
\usepackage{multirow}
\usepackage{graphicx}
\usepackage{amsmath}
\usepackage{amssymb}

\usepackage{array}
\newcolumntype{C}[1]{>{\centering\arraybackslash}p{#1}}
\definecolor{customlightblue}{rgb}{0.25, 0.25, 1.0}
\definecolor{custompink}{rgb}{1.0, 0.1, 0.6} 

\newcommand\method{\textsc{Deer}}
\newcommand\kate{\textsc{Kate}}

\title{LLMs are Better Than You Think: Label-Guided In-Context Learning for Named Entity Recognition}

\author{Fan Bai$^{\clubsuit}$ Hamid Hassanzadeh$^{\diamondsuit}$  Ardavan Saeedi$^{\diamondsuit}$ \textbf{Mark Dredze}$^{\clubsuit}$ \\
\textsuperscript{$\clubsuit$}\text{Johns Hopkins University}, \textsuperscript{$\diamondsuit$}\text{Optum} \\
 \small \texttt{fbai3@jh.edu}, \texttt{\{hamid.hassanzadeh, ardavan.saeedi\}@optum.com}, \texttt{mdredze@cs.jhu.edu} \\
 \\
 }

\begin{document}
\maketitle
\begin{abstract}

In-context learning (ICL) enables large language models (LLMs) to perform new tasks using only a few demonstrations. However, in Named Entity Recognition (NER), existing ICL methods typically rely on task-agnostic semantic similarity for demonstration retrieval, which often yields less relevant examples and leads to inferior results. We introduce \textbf{\method{}}, a training-free ICL approach that enables LLMs to make more informed entity predictions through the use of label-grounded statistics. \method{} leverages token-level statistics from training labels to identify tokens most informative for entity recognition, enabling entity-focused demonstrations. It further uses these statistics to detect and refine error-prone tokens through a targeted reflection step. Evaluated on five NER datasets across four LLMs, \method{} consistently outperforms existing ICL methods and achieves performance comparable to supervised fine-tuning. Further analyses demonstrate that \method{} improves example retrieval, remains effective on both seen and unseen entities, and exhibits strong robustness in low-resource settings.\footnote{Code and data are available at \url{https://github.com/bflashcp3f/deer}.}

\end{abstract}

\section{Introduction}
\label{sec:intro}

Identifying and categorizing named entities in text (Named Entity Recognition, NER) is a core information extraction (IE) task with applications in diverse domains including science \citep{lin2004maximum, leser2005makes, luan-etal-2018-multi, bai-etal-2022-synkb}, news \citep{tjong-kim-sang-de-meulder-2003-introduction, whitelaw2008web}, and social media \citep{finin-etal-2010-annotating, ritter-etal-2011-named,li2012twiner}. Supervised learning works well for NER using models with tailored architectures, fine-tuned on large labeled datasets \citep{nadeau2007survey, ma-hovy-2016-end}. However, these approaches do not generalize well across domains or to new entity types without re-training on annotated data, thereby constraining their broader applicability \citep{lin-lu-2018-neural, bai-etal-2021-pre}.

\begin{figure}[!t]
\begin{center}
  \includegraphics[width=0.5\textwidth]
  {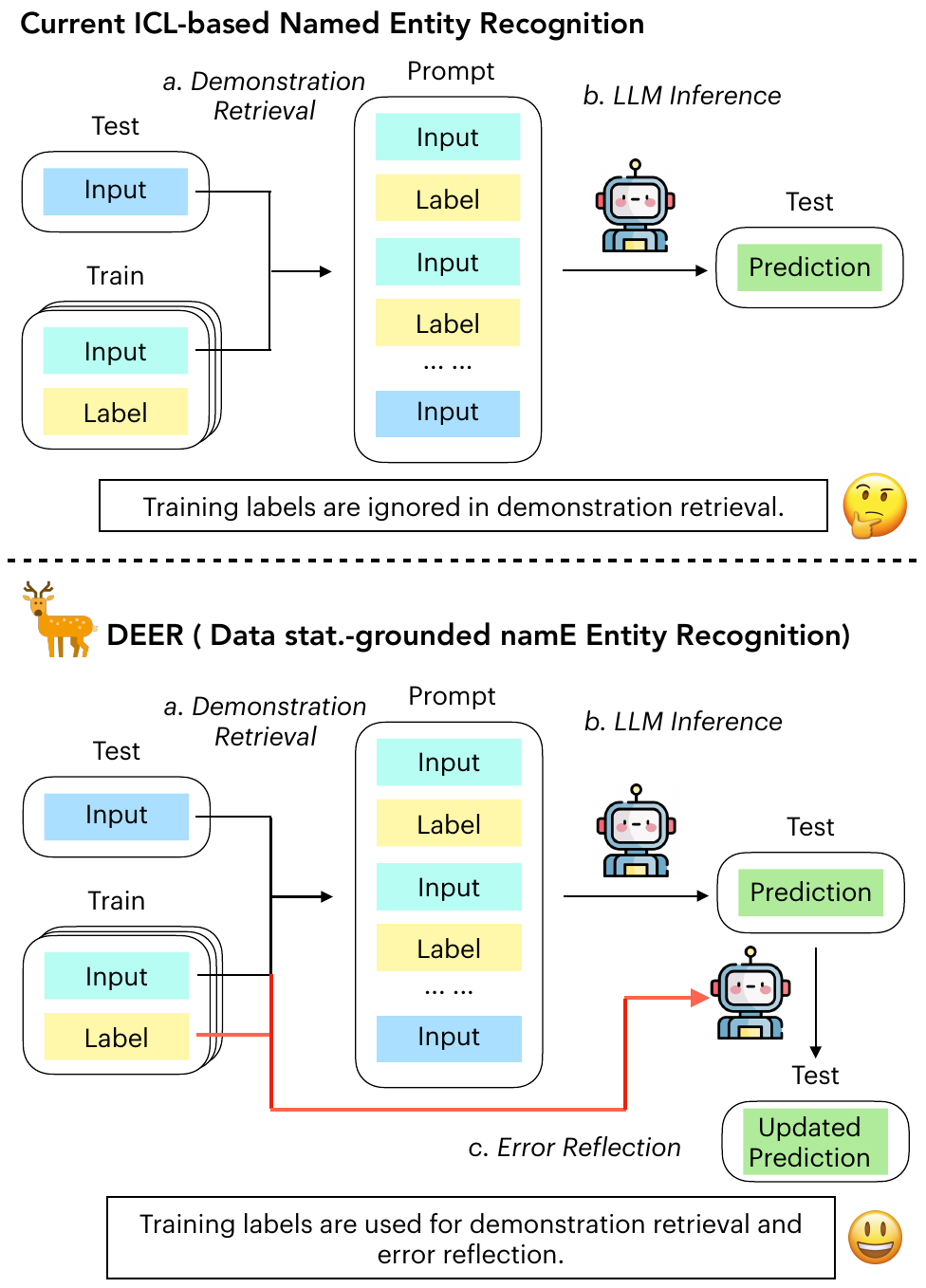}
\caption{Existing ICL methods for NER typically select demonstrations based on task-agnostic input embeddings, which we argue yield suboptimal demonstrations for LLMs and consequently inferior performance. In contrast, our proposed method, \method{}, leverages label-grounded statistics to identify entity-relevant tokens, enabling more task-focused demonstration selection and targeted error reflection.
}
  \label{fig:motivation}
\end{center}
\end{figure}

\begin{figure*}[ht!]
    \centering
    \includegraphics[width=\textwidth]{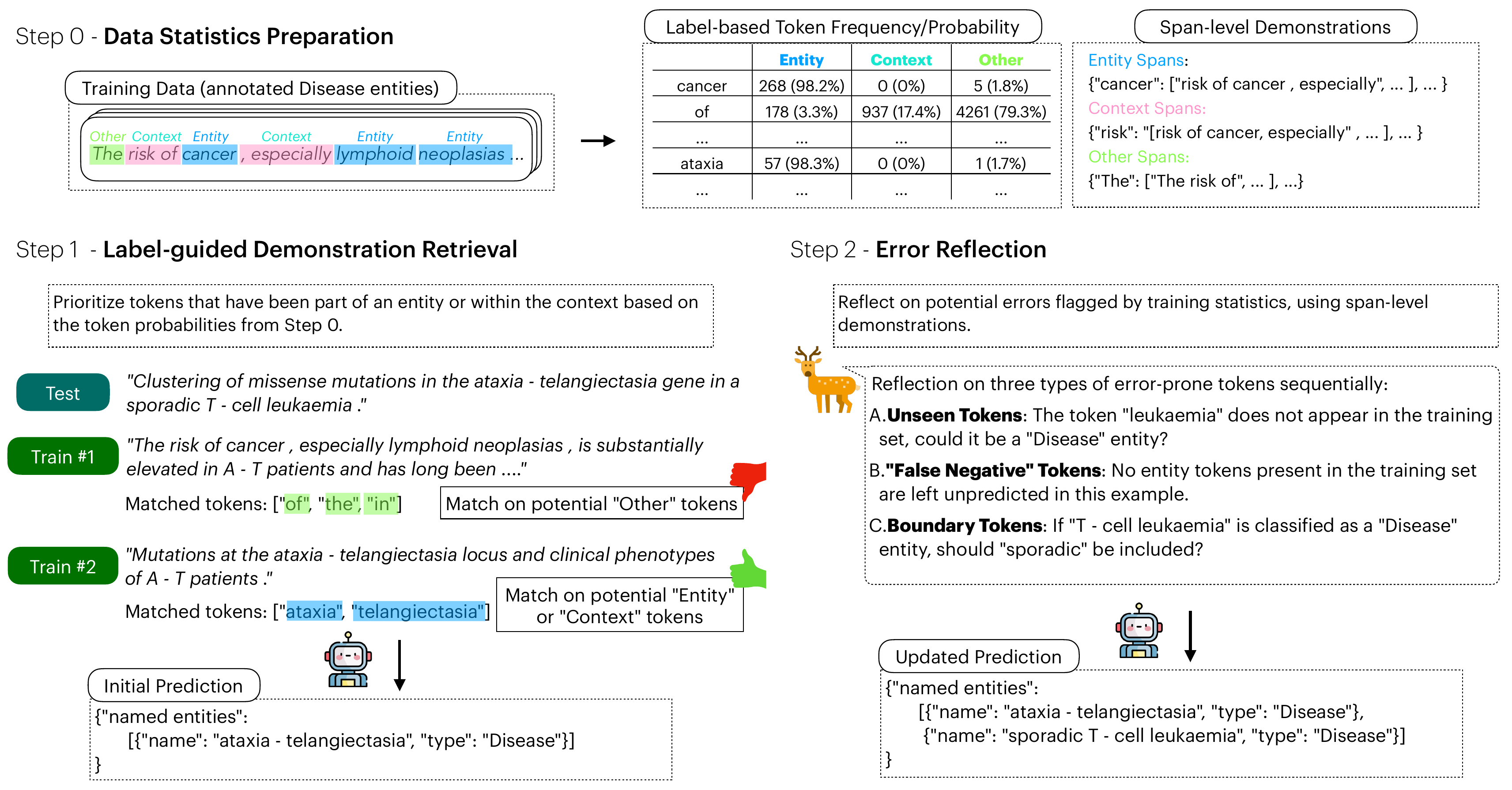}
    \caption{Overview of \method{}. In the preparation stage (Step 0), the method compiles training input and labels to compute token frequencies and probabilities in three scenarios: 1) \textcolor{customlightblue}{entity token}, 2) \textcolor{custompink}{context token}, and 3) \textcolor{green}{other token}, along with their associated spans. In the inference stage (Step 1 and 2), Step 1 retrieves sentence-level demonstrations by emphasizing potential entity- and context-related tokens based on probabilities from Step 0. Step 2 refines predictions from Step 1 by addressing error-prone tokens based on label statistics, focusing on three token types: unseen tokens, "false negative" tokens, and boundary tokens. For each token type, the refinement process retrieves span-level demonstrations and prompts LLMs to adjust predictions. See \S\ref{sec:method} for further details.
    }
    \label{fig:method}
\end{figure*}

Recent advances in large language models (LLMs; \citealp{brown2020languagemodelsfewshotlearners, touvron2023llamaopenefficientfoundation, dubey2024llama3herdmodels, openai2024gpt4ocard}) suggest an alternative: \textit{in-context learning} (ICL; \citealp{dong2024surveyincontextlearning}), where models learn new tasks directly from examples provided as input, without parameter updates. As shown in Figure~\ref{fig:motivation}, a common ICL strategy for NER \citep{wang2023gptnernamedentityrecognition, lu2024largelanguagemodelsstruggle} involves retrieving a fixed number of demonstrations using embeddings of input data, and then prompting the model to extract entities. While such methods show particular promise in low-resource scenarios \citep{lee-etal-2022-good}, it does not scale as efficiently as supervised methods \citep{sainz2024gollieannotationguidelinesimprove}. For example, \citet{monajatipoor2024llmsbiomedicinestudyclinical} report a 79.3 F\textsubscript{1} score on the NCBI dataset using GPT-4 with 16 examples per query. In contrast, a fine-tuned BERT-base model achieves an 84.0 F\textsubscript{1} score when given access to the same training data. 

In this paper, we argue that this performance gap primarily stems from inefficiencies in demonstration retrieval rather than from inherent limitations of LLMs. The prevailing sentence embedding–based retrievers are ill-suited for NER for two key reasons: (1) NER is a token-level task requiring fine-grained sensitivity to entity boundaries, whereas sentence embeddings operate at a coarse, sentence-level granularity; and (2) these embeddings are task-agnostic, disregarding label information from the training data, even though an ideal retriever should recognize entity-focused patterns and retrieve demonstrations containing similar entities to those in the query. As shown in Table~\ref{tab:retriever_analysis}, our sanity-check analysis confirms that sentence embedding–based retrievers frequently fail to retrieve demonstrations containing overlapping entities with the query, highlighting their inefficiency.

Motivated by these findings, we propose \method{} (\textbf{\textsc{D}}ata Stat.-grounded nam\textbf{\textsc{E}}d \textbf{\textsc{E}}ntity \textbf{\textsc{R}}ecognition), a method designed to help LLMs make more informed predictions during ICL in a training-free manner. The core idea of \method{} is to leverage task-specific information embedded in label statistics. In the context of NER, this corresponds to identifying which tokens are more likely to appear within entity spans, near entity spans, or outside them. Based on such statistics from the entire training set, \method{} incorporates them in two complementary stages. First, the retriever uses token-level statistics to guide demonstration selection, prioritizing examples containing tokens that are likely to correspond to entities or their contexts, resulting in more relevant demonstrations. Second, during an error reflection stage, \method{} revisits potentially misclassified tokens (flagged by these statistics) with targeted span-level demonstrations, enabling the model to refine its predictions and achieve improved overall performance.

We evaluate \method{} on five NER datasets spanning four domains: news, biomedicine, social media, and mixed web text, using four LLMs: \texttt{Qwen2.5-7B}, \texttt{Llama3.3-70B}, \texttt{GPT-4o-mini}, and \texttt{GPT-4o}. The results show that \method{} consistently outperforms ICL baselines across datasets and models. With \texttt{GPT-4o}, \method{} achieves performance competitive with supervised models on four out of five datasets, while \texttt{Llama3.3-70B} also demonstrates strong results. Further analysis confirms that \method{} enhances example retrieval, improves recognition of both seen and unseen entities, and demonstrates robustness in low-resource settings.

\section{\method{} \includegraphics[scale=0.04]{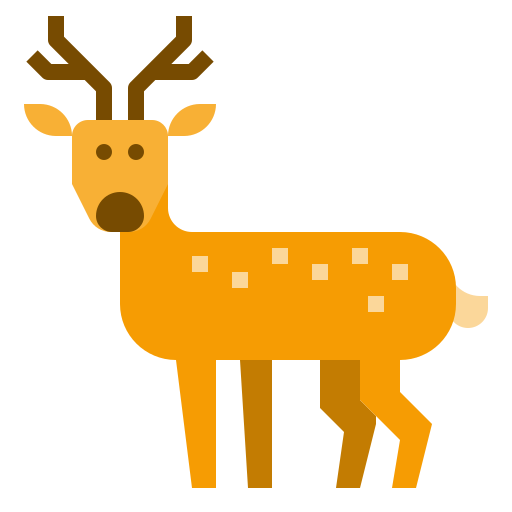} }
\label{sec:method}

We introduce \textbf{\method{}}, an ICL approach that enables LLMs to make more informed entity predictions through the use of label statistics. As illustrated in Figure~\ref{fig:method}, \method{} consists of two main components: (1) label-guided retrieval and (2) error reflection. The following sections describe the problem setup and the construction of the label statistics in \method{}, followed by explanations of each component.

\subsection{Problem Setup}

Named entity recognition aims to identify all text spans corresponding to pre-defined entity types in a sentence. Following prior work on LLM-based entity recognition \citep{dunn2022structuredinformationextractioncomplex, lu-etal-2022-unified}, we formulate the task as a span identification problem and format the output as a structured JSON object (Figure~\ref{fig:motivation}), using the key \textit{``named\_entities''} for a list of JSON objects, each containing the name and type of an entity (e.g., \textit{\{``name'': ``Barack Obama'', ``type'': ``PERSON''\}}). In Appendix~\ref{sec:ablation_appendix}, we compare it with two other prompting formats from recent work and find that our chosen format yields the best performance.

\subsection{Data Statistics Preparation}
As motivated in \S\ref{sec:intro}, an effective demonstration retriever should capture entity-focused patterns and retrieve demonstrations that exhibit similar patterns to those in the query. To achieve this, we leverage label-grounded token statistics from the training data. Specifically, we categorize all training tokens into three classes based on their relation to entities and compute the frequency and probability of each token within these classes (Step~0 in Figure~\ref{fig:method}).

\textcolor{customlightblue}{Entity tokens} are those that belong to an identified entity. \textcolor{custompink}{Context tokens} are the tokens surrounding an entity, typically two tokens on each side, which provide identification cues and help define entity boundaries \citep{huang2015bidirectionallstmcrfmodelssequence}. For example, consider the sentence in Figure \ref{fig:method}, \texttt{``The risk of cancer, ...''}, where ``cancer'' is an \textcolor{customlightblue}{entity token}, and ``risk'' and ``of'' are \textcolor{custompink}{context tokens}. \textcolor{custompink}{Context tokens} like ``risk of'' provide contextual clues for identifying Disease entities, even in cases where entities are unseen in the training data. Finally, \textcolor{green}{\textbf{other tokens}} include all remaining tokens that fall outside the entity and its immediate context window (e.g., ``The'' in the example).

Moreover, we define token-focused spans for these categories: \textcolor{customlightblue}{entity spans} include the entity along with its surrounding context, \textcolor{custompink}{context spans} are similar but correspond to context tokens, and \textcolor{green}{other spans} consist of two adjacent tokens on each side of an \textcolor{green}{other token}. For instance, the span \texttt{``risk of cancer, especially''} serves as an \textcolor{customlightblue}{entity span} for ``cancer'' but as a \textcolor{custompink}{context span} for ``risk.''\footnote{In our implementation, \textcolor{customlightblue}{entity spans} exclude neighboring entities within the context window to encourage greater attention on context tokens for more generalizable entity recognition. For example, given the sentence ``Lionel Messi and Cristiano Ronaldo are exceptional football players'' and a context window of two, the \textcolor{customlightblue}{entity spans} for the two PERSON entities are \texttt{``Lionel Messi and''} and \texttt{``and Cristiano Ronaldo are exceptional''}.} These label-dependent spans offer a more precise and structured approach to handling tokens in different scenarios. In \S\ref{sec:ablation}, we analyze the impact of incorporating these three token types and demonstrate the importance of \textcolor{custompink}{context tokens}, particularly for improving performance on unseen entities.

\subsection{Label-guided Retrieval} 

\method{} begins with label-guided, token-based retrieval for ICL (Step 1 in Figure \ref{fig:method}). The ICL prompt consists of the task description, specifying the targeted entity types and output format, followed by $N$ examples selected from the training set and the query sentence. Unlike previous work that uses task-agnostic sentence embeddings, we aim to address the token-centric nature of NER. Specifically, to compute the similarity between a training sentence \( s_i \) and a test sentence \( s_t \), we define a similarity score $\mathbf{S}_{s_{i}, s_{t}}$ that combines exact token matching and embedding-based semantic similarity:

\small
\begin{equation}
\mathbf{S}_{s_{i}, s_{t}} = \lambda_1 \cdot S_{\text{token}}(s_i, s_t) + \lambda_2 \cdot S_{\text{embed}}(s_i, s_t),
\end{equation}
\normalsize
where \( \lambda_1 \) and \( \lambda_2 \) are hyper-parameters controlling the relative contribution of two components. The token match score \( S_{\text{token}} \) is computed as:

\small
\begin{equation}
S_{\text{token}}(s_i, s_t) = \sum_{t \in s_t} \mathbb{I}(t \in s_i) \cdot W(t),
\end{equation}
\normalsize
where \( \mathbb{I}(t \in s_i) \) is an indicator function that equals 1 if test token \( t \) appears in the training sentence \( s_i \), and 0 otherwise. The function \( W(t) \) assigns importance to each token:

\scriptsize
\begin{equation}
W(t) =
\begin{cases} 
w_{e} P(t_e) + w_{c} P(t_c) + w_{o} P(t_o), & \text{if } t \text{ is seen}, \\
1, & \text{otherwise }
\end{cases}
\end{equation}
\normalsize
\noindent where \( P(t_e) \), \( P(t_c) \), and \( P(t_o) \) are the probabilities of token \( t \) being an \textcolor{customlightblue}{entity}, \textcolor{custompink}{context}, or \textcolor{green}{other token}, respectively, in the training corpus. The hyperparameters \( w_e \), \( w_c \), and \( w_o \) control the relative importance of these token types, allowing the model to prioritize \textcolor{customlightblue}{entity} and \textcolor{custompink}{context} tokens, which are more informative for NER.\footnote{For all hyperparameters, we perform a grid search over three possible values: $[1.0, 0.5, 0.01]$, and find that setting \( \lambda_1 = 1.0 \), \( \lambda_2 = 1.0 \), \( w_{e} = 1.0 \), \( w_{c} = 1.0 \), and \( w_{o} = 0.01 \) performs well in most cases. See ablations in Appendix~\ref{sec:ablation_appendix}.} For semantic similarity, we compute a sentence embedding by aggregating token embeddings:
\begin{equation}
\mathbf{v}_{s} = \sum_{t \in s} W(t) \cdot \mathbf{v}_t,
\end{equation}
where \( \mathbf{v}_t \) represents the embedding of token \( t \). The final score is computed via cosine similarity:
\begin{equation}
S_{\text{embed}}(s_i, s_t) = \cos(\mathbf{v}_{s_i}, \mathbf{v}_{s_t}).
\end{equation}
By integrating both token matching and embedding-based similarity, our method ensures precise retrieval of seen tokens while also accounting for unseen tokens, leading to optimized performance across different dataset characteristics.

\subsection{Error Reflection (ER)}

Even with improved retrieval, ICL can only incorporate a small subset of training examples, leaving blind spots for LLM predictions. For example, the model may miss entities present in the training set but absent from retrieved examples or make boundary errors due to insufficient guidance. To address this, we introduce an error reflection step that uses prepared statistics to identify error-prone tokens and provides targeted demonstrations to help the model revise its predictions. We focus on single-entity, boundary-related errors, which have been identified as a major source of NER mistakes \citep{ding-etal-2024-rethinking, lu2024largelanguagemodelsstruggle}. A hierarchical breakdown of NER errors is shown in Figure~\ref{fig:error_onto}.

As shown in Step 2 of Figure~\ref{fig:method}, our reflection step targets three token categories associated with potential errors: unseen tokens, “false negative” tokens, and boundary tokens of predicted entities. The first two categories help recover missed entities, while the third addresses both false positives and boundary-related errors. The reflection process proceeds sequentially. We first examine unseen and false negative tokens in the query sentence to identify additional candidate entities. Then, we refine or discard predictions by inspecting boundary tokens. To enable this process, we use a chain-of-thought reasoning approach \citep{wei2023chainofthoughtpromptingelicitsreasoning}, prompting the model to reason about the relationship between the query and retrieved demonstrations before making final predictions. Details on how we identify these tokens and construct reflection prompts for each category are provided below. Exact prompts are listed in Appendix~\ref{sec:full_prompt}.

\paragraph{Unseen Tokens} 
Unseen tokens, defined as tokens not present in the training data, often lead to false negatives since they lack direct matches for retrieval. To address this, we leverage their surrounding context. Specifically, we select unseen tokens that were not predicted as part of an entity but are surrounded by tokens frequently labeled as entities or context tokens in the training set. This targeted strategy avoids reflecting on every unseen token, reducing inference cost while improving prediction accuracy (see Table~\ref{tab:reflect_freq} in the appendix for reflection frequency statistics). For each selected unseen token, we retrieve $M$ span-level demonstrations from the training set by matching its context tokens. A chain-of-thought prompt then guides the model to reason through these demonstrations and determine whether a new entity should be extracted. For sentences with multiple unseen tokens, this process is applied in the order the tokens appear.

\paragraph{``False Negative" Tokens} 
These are tokens commonly annotated as entities in the training set but missed during initial predictions. A token is flagged as a ``false negative" if its entity likelihood in the training set exceeds a predefined threshold. For each such token, we retrieve $M$ span-level examples containing it and prompt the model to predict whether an entity should be extracted.

\paragraph{Boundary Tokens}
These tokens lie at the edges of predicted entities and are critical for refining entity spans or eliminating false positives. For each prediction, we examine $2K$ tokens around the boundary: $K$ edge tokens of the entity and $K$ from the surrounding context. We focus on tokens likely to be misclassified based on training statistics, for example, tokens predicted as part of an \textcolor{customlightblue}{entity} but appeared more as \textcolor{custompink}{context tokens} in the training set. For each reflected token, we retrieve span-level examples from all three categories: \textcolor{customlightblue}{entity}, \textcolor{custompink}{context}, and \textcolor{green}{other}. This variety allows the model to reason through different scenarios and decide whether to include each token, ultimately refining the entity's boundaries.

\section{Experimental Setup}
\label{sec:experiment}

\subsection{Dataset and Evaluation Metric}

To demonstrate the effectiveness of our method, we evaluate it on five NER datasets spanning four domains, which have been used in recent ICL-based NER studies \citep{wang2023gptnernamedentityrecognition, monajatipoor2024llmsbiomedicinestudyclinical}. These datasets include CoNLL03 \citep{tjong-kim-sang-de-meulder-2003-introduction} for news, NCBI-disease \citep{dougan2014ncbi} and bc2gm \citep{smith2008overview} for biomedicine, TweetNER7 \citep{ushio-etal-2022-named} for social media, and OntoNotes \citep{weischedel2013ontonotes} for web text from various sources. Dataset statistics are provided in Table \ref{tab:data_stat}. 
We refer readers to the corresponding papers for details on each dataset. Following prior work \citep{berger2024incontextlearningbudgetcase, bai-etal-2024-schema}, we control computational costs by limiting the test set size to 1,000 examples through random sampling of their original test sets. To ensure the sampled subset provides a reliable estimate of model performance, we calculate 95\% confidence intervals using 1,000 bootstrap resamples, and the margins of error fall within a 0.01-0.03 range, verifying the validity of our performance estimates. Further details on this process can be found in Appendix \ref{sec:stat_test}.

All experiments are evaluated using the standard mention-level matching metric for NER \citep{tjong-kim-sang-de-meulder-2003-introduction}, where a predicted entity is considered correct only if it matches both the span boundary and the entity type.

\begin{table}[h!]
\small
\begin{center}
\scalebox{0.85}{
\begin{tabular}{lccc}
\toprule
 & & & \textbf{\# Sentences} \\
\textbf{Dataset} & \textbf{Domain} & \textbf{\# Ent.} & (Tr / Dv / Ts) \\
\midrule

CoNLL03 \citeyearpar{tjong-kim-sang-de-meulder-2003-introduction} & News & 4 & 14.0k / 3.3k / 3.5k \\
OntoNotes \citeyearpar{weischedel2013ontonotes} & Web & 18 & 60.0k / 8.5k / 8.3k \\
NCBI \citeyearpar{dougan2014ncbi} & Biomed & 1 & 5.4k / 0.9k / 0.9k \\
BC2GM \citeyearpar{smith2008overview} & Biomed & 1 & 12.5k / 2.5k / 5.0k \\
TweetNER7 \citeyearpar{ushio-etal-2022-named} & Social & 7 & 7.1k / 0.8k / 0.6k \\
\bottomrule
\end{tabular}
}
\end{center}
\caption{\label{tab:data_stat} Statistics of the five experimented datasets. Each dataset contains a minimum of 5,000 annotated training examples, ensuring competitive performance of fine-tuned models.
}
\end{table}

\begin{table*}[!ht]
\centering
\resizebox{0.85\textwidth}{!}{
\begin{tabular}{clC{1.5cm}C{1.5cm}C{1.5cm}C{1.5cm}C{1.5cm}C{1.5cm}C{1.5cm}}
    \toprule
    
    \textbf{LLM} & \textbf{Method} & NCBI & bc2gm & CoNLL03 & OntoNotes &  TwNER7 & average \\
    \midrule
    \multirow{5}{*}{\texttt{Qwen2.5-7B}}   & BM25 & 66.8 & 57.0 & 83.0 & 62.6 & 55.0 & 64.9  \\
                                           & BSR \citeyearpar{gupta-etal-2023-coverage} & 68.0 & 59.9 & 86.0 & 66.8 & 56.1 & 67.4 \\
                                           & \kate{} \citeyearpar{liu-etal-2022-makes} & 66.9 & 59.3 & 83.7 & 63.8 & 57.1 & 66.2 \\
                                           & DEER\textsubscript{w/o ER} & 71.7 & 62.5 & 87.2 & 68.4 & 57.3 & 69.4 \\
                                           & DEER & \textbf{73.2} & \textbf{63.6} & \textbf{87.8} & \textbf{71.8} & \textbf{57.5} & \textbf{70.8} \\
    \midrule
    \multirow{5}{*}{\texttt{Llama3.3-70B}} & BM25 & 79.7 & 65.5 & 88.9 & 76.0 & 61.7 & 74.4  \\
                                           & BSR \citeyearpar{gupta-etal-2023-coverage} & 79.3 & 67.7 & 90.2 & 78.9 & 62.8 & 75.8 \\
                                           & \kate{} \citeyearpar{liu-etal-2022-makes} & 79.4 & 66.5 & 88.3 & 76.1 & 62.0 & 74.5 \\
                                           & DEER\textsubscript{w/o ER} & 81.9 & 68.6 & 89.6 & 78.9 & 63.3 & 76.5 \\
                                           & DEER & \textbf{83.9} & \textbf{69.6} & \textbf{90.2} & \textbf{80.2} & \textbf{63.5} & \textbf{77.5} \\
    \midrule
    \multirow{5}{*}{\texttt{GPT-4o-mini}}  & BM25 & 74.1 & 62.9 & 87.8 & 74.4 & 59.3 & 71.7  \\
                                           & BSR \citeyearpar{gupta-etal-2023-coverage} & 74.9 & 62.9 & 89.6 & 77.4 & 61.1 & 73.2 \\
                                           & \kate{} \citeyearpar{liu-etal-2022-makes} & 76.0 & 64.1 & 89.2 & 74.9 & 59.6 & 72.8 \\
                                           & DEER\textsubscript{w/o ER} & 77.3 & 65.4 & 89.9 & 77.0 & 60.4 & 74.1 \\
                                           & DEER & \textbf{79.3} & \textbf{67.1} & \textbf{90.9} & \textbf{79.2} & \textbf{61.6} & \textbf{75.8} \\
    \midrule
    \multirow{5}{*}{\texttt{GPT-4o}}       & BM25 & 81.0 & 71.9 & 92.0 & 81.2 & 62.8 & 77.8   \\
                                           & BSR \citeyearpar{gupta-etal-2023-coverage} & 81.2 & 73.5 & 93.3 & 82.9 & 63.9 & 78.9 \\
                                           & \kate{} \citeyearpar{liu-etal-2022-makes} & 80.0 & 73.6 & 92.9 & 80.7 & 62.2 & 77.9 \\
                                           & DEER\textsubscript{w/o ER} & 82.7 & 74.3 & 93.5 & 83.5 & 63.6 & 79.5 \\
                                           & DEER & \textbf{84.8} & \textbf{75.4} & \textbf{93.6} & \textbf{85.2} & \textbf{64.3} & \textbf{80.7} \\
\bottomrule
\end{tabular}
}

\caption{Comparison of \method{} with three ICL baselines across five datasets using four recent LLMs and 8 demonstrations per query. \method{}\textsubscript{w/o ER} excludes the error reflection step and already outperforms all baselines, validating the effectiveness of our label-guided retrieval. Adding error reflection yields further improvements. Among the LLMs, \texttt{GPT-4o} performs best overall, while \texttt{Llama3.3-70B} surpasses \texttt{GPT-4o-mini} and approaches \texttt{GPT-4o}, demonstrating the promise of open-source models.
}
\label{tab:icl_results}
\end{table*}

\begin{table}[h!]
\small
\begin{center}
\scalebox{0.8}{
\begin{tabular}{lcccccc}
\toprule
 & & & \multicolumn{4}{c}{\textbf{Unseen Entity}} \\
   \cmidrule(l){4-7} 
 & \multicolumn{2}{c}{\textbf{Seen Entity}} & \multicolumn{2}{c}{Seen Token} & \multicolumn{2}{c}{Unseen Token} \\
 \cmidrule(l){2-3} \cmidrule(l){4-5} \cmidrule(l){6-7}
 & \kate{} & \method{} & \kate{} & \method{} & \kate{} & \method{} \\
\midrule
NCBI & 88.7 & \textbf{91.4} & 61.1 & \textbf{66.5} & 70.3 & \textbf{81.0} \\
bc2gm & 85.8 & \textbf{88.4} & \textbf{60.5} & 60.1 & 70.8 & \textbf{75.0} \\
CoNLL03 & 95.5 & \textbf{95.7} & 84.9 & \textbf{87.7} & 91.2 & \textbf{92.1} \\
OntoNotes  & 86.0 & \textbf{89.6} & 65.5 & \textbf{72.4} & 76.1 & \textbf{79.7} \\
TweetNER7  & 71.1 & \textbf{74.9} & 45.9 & \textbf{46.9} & 58.3 & \textbf{58.3} \\
\bottomrule
\end{tabular}
}
\end{center}
\caption{\label{tab:unseen} Breakdown of test set performance for \kate{} and \method{} (with \texttt{GPT-4o}) based on entity presence in the training data. Unseen entities are further categorized by whether they contain novel tokens. \method{} outperforms \kate{} on both seen and unseen entities, highlighting its effectiveness to novel entities.}
\end{table}

\begin{table*}[!ht]
\centering
\resizebox{1.0\textwidth}{!}{
\begin{tabular}{lcccccccccccc}
    \toprule
    
    &  \multicolumn{2}{c}{8 demo.} & \multicolumn{2}{c}{32 demo.} & \multicolumn{5}{c}{\textbf{Fine-tuned}}  \\

    \cmidrule(l){2-3}  \cmidrule(l){4-5}  \cmidrule(l){6-10} 
    
    \textbf{Dataset} & \method{}\textsubscript{w/o ER} &  \method{} & \method{}\textsubscript{w/o ER} & \method{} & BERT & UniNER\textsubscript{7B} & InsUIE\textsubscript{11B} & GoLLIE\textsubscript{13B} & GoLLIE\textsubscript{34B} \\
    \midrule

    NCBI  & 82.7 (\$2.1) & 84.8 (\$3.0) & 84.1 (\$6.5) & 85.7 (\$7.3) & 84.0 & 87.0 & 86.2 & 86.5 & 85.8 \\
    bc2gm  & 74.3 (\$2.9) & 75.4 (\$3.7) & 75.0 (\$9.4) & 75.6 (\$9.9) & 79.0 & 82.4 & 80.7	& - & - \\
    CoNLL03 & 93.5 (\$1.9) & 93.6 (\$2.2) & 93.9 (\$5.8) & 94.0 (\$6.0) & 90.4 & 93.3 & 91.5 & 93.0 & 93.1 \\
    OntoNotes & 83.5 (\$2.5) & 85.2 (\$3.3) & 85.8 (\$7.5) & 86.6 (\$8.2) & 87.2 & 89.9 & 88.7 & 84.0 & 84.6\\
    TweetNER7 & 63.6 (\$1.9) & 64.3 (\$3.4) & 64.9 (\$6.1) & 65.3 (\$7.2) & 60.9 & 65.7 & 66.0 & - & - \\

\bottomrule
\end{tabular}
}
\caption{ICL results of \method{} (using \texttt{GPT-4o}) with 8 and 32 demonstrations, compared to five fine-tuning baselines. Values in parentheses indicate associated API inference costs on the test set. Increasing the number of demonstrations improves performance, and error reflection continues to provide additional gains at larger scales. Notably, \method{} with 8-shot error reflection performs competitively with the 32-shot setting without reflection, but at a significantly lower cost. \method{} also matches or exceeds fine-tuning baselines on four out of five datasets.}
\label{tab:scaling_demo_ft_results}
\end{table*}

\subsection{Baselines} 

We compare \method{} against both ICL and fine-tuning baselines to demonstrate its improvements over standard ICL methods and its ability to approach fine-tuning performance. All methods use the full training set for either fine-tuning or in-context demonstration selection.

\paragraph{ICL Baselines} We cover three ICL baselines: KATE \citep{liu-etal-2022-makes}, BM25 \citep{robertson2009probabilistic}, and BSR \citep{gupta-etal-2023-coverage}. \textbf{\kate{}} serves as our primary ICL baseline, as it remains the dominant framework for ICL-based NER \citep{wang2023gptnernamedentityrecognition, berger2024incontextlearningbudgetcase, monajatipoor2024llmsbiomedicinestudyclinical}. This method selects in-context examples through nearest neighbor search using sentence-level embeddings. \textbf{BM25} treats both the test input and demonstrations as bags of words and measures their relevance via a TF-IDF–weighted recall of overlapping words. Recent IE work \citep{sun-etal-2024-leveraging} has shown its effectiveness for demonstration retrieval. \textbf{BSR} employs BERTScore-Recall \citep{BERTScore} as a similarity metric, which measures token-level recall using token embeddings. While mainly used in semantic parsing, we adapt it here for NER. To ensure the competitiveness of the baselines, we evaluate seven embedding models commonly used in recent studies (see Appendix~\ref{sec:ablation_appendix}). OpenAI’s \texttt{text-embedding-3-small} consistently outperforms open-source alternatives such as \texttt{all-mpnet-base-v2}, and is therefore used in all our experiments.
We do not include \textbf{GPT-NER} \citep{wang2023gptnernamedentityrecognition}, an entity-based retriever, due to its impracticality for the ICL setting. Nonetheless, we show in Table~\ref{tab:gpt_ner} that our method outperforms GPT-NER. Additional discussion on GPT-NER can be found in \S\ref{sec:related_work}.

\paragraph{Fine-tuning Baselines} 

We compare with five supervised models: BERT-base, UniversalNER-7B \citep{zhou2024universalner}, InstructUIE-11B \citep{wang2023instructuie}, and GoLLIE-13B/34B \citep{sainz2024gollieannotationguidelinesimprove}.  \textbf{BERT-base} \citep{devlin-etal-2019-bert} is fine-tuned for token-level classification; we run three random seeds and report the averaged performance. \textbf{UniversalNER-7B} applies targeted distillation for NER by prompting \texttt{gpt-3.5-turbo} to annotate unlabeled text, training a student model on the refined data, and subsequently fine-tuning it on supervised NER datasets.\footnote{As noted in its \href{https://github.com/universal-ner/universal-ner}{GitHub repository}, the authors have not released the code or prompts necessary to reproduce most supervised results due to licensing restrictions. We therefore rely on the reported results in the original paper.} \textbf{InstructUIE-11B} adopts unified instruction tuning, training FlanT5-11B \citep{chung2022scalinginstructionfinetunedlanguagemodels} on 32 diverse IE datasets reformulated into a text-to-text format with expert-written instructions, yielding strong performance in both supervised and low-resource settings. Finally, \textbf{GoLLIE-13B} and \textbf{GoLLIE-34B} extend CodeLlama \citep{rozière2024codellamaopenfoundation} with human-authored guidelines, leading to substantial improvements in zero-shot IE while also maintaining competitive supervised results.

\subsection{Implementation Details} 

We conduct experiments with four recent LLMs, covering both open-source and proprietary models: \texttt{Qwen2.5-7B} \citep{qwen2025qwen25technicalreport}, \texttt{Llama3.3-70B} \citep{dubey2024llama3herdmodels}, \texttt{GPT-4o-mini}, and \texttt{GPT-4o} \citep{openai2024gpt4ocard}.\footnote{\texttt{gpt-4o-mini-2024-07-18} and \texttt{gpt-4o-2024-08-06} are used through OpenAI API. In our preliminary experiments, we also evaluated \texttt{Llama3.1-8B}. However, it failed to generate outputs in the required format during ICL (see Appendix~\ref{sec:llama_issue} for more details). Thus, we did not pursue it further.}
We use the same prompt across four LLMs and set the decoding temperature to 0 to facilitate result reproducibility. For main ICL experiments, we retrieve eight demonstrations per query, sorted in ascending order of similarity to the test example. 
Appendix \ref{sec:implement_details} contains detailed descriptions of the prompts and all hyperparameters.

\section{Results}
\label{sec:results}

\paragraph{Comparison with ICL Baselines}  
Table~\ref{tab:icl_results} presents the results of three ICL baselines and two variants of \method{}, including the one without error reflection (\method{}\textsubscript{w/o ER}), evaluated across five datasets and four LLMs. \method{}\textsubscript{w/o ER} consistently outperforms \kate{} in all settings, demonstrating the effectiveness of our label-guided, token-level demonstration retrieval. It also performs better than BM25 and BSR. Interestingly, BSR consistently outperforms \kate{}, suggesting that BERTScore-Recall, which emphasizes token-level coverage, may be more suitable than cosine similarity on sentence representations for token-centric tasks like NER. A promising future direction would be to integrate our label-guided retrieval with BERTScore-Recall. Adding error reflection leads to further gains across all settings, confirming its effectiveness. A breakdown of performance across the three reflection steps is provided in Table~\ref{tab:breakdown} in the appendix. Appendix \ref{sec:error} (Table \ref{tab:error_breakdown_full}) presents an error analysis, highlighting \method{}'s advantages in reducing under-predicted entities and boundary errors. 

Among the LLMs, \texttt{GPT-4o} achieves the best overall results while our method delivers larger relative improvements on smaller models, such as \texttt{Qwen2.5-7B}. For example, on OntoNotes, \method{} achieves an 8.0 absolute F\textsubscript{1} gain over \kate{}. \texttt{Llama3.3-70B} shows strong performance, outperforming \texttt{GPT-4o-mini} and approaching \texttt{GPT-4o}, highlighting the potential of open-source LLMs.

Table \ref{tab:unseen} further analyzes performance based on entity presence in the training data. The test set entities are categorized as either seen or unseen during training. Unseen entities are further divided into those containing novel tokens and those formed by permutations of seen tokens. The results show that \method{} consistently outperforms \kate{} on both seen and unseen entities, demonstrating its effectiveness in generalizing to novel entities.

\paragraph{Scaling Demonstrations and Comparison with Fine-tuning Baselines}  
To evaluate the scalability of \method{} and its performance relative to fine-tuning, Table~\ref{tab:scaling_demo_ft_results} reports results using \texttt{GPT-4o} with 8 and 32 demonstrations per query, along with five fine-tuning baselines. We include both performance and API inference costs, with and without error reflection.\footnote{Inference costs are calculated using OpenAI's \href{https://openai.com/api/pricing/}{official pricing}: \$2.50 per million input tokens and \$10.00 per million output tokens.} Increasing the number of demonstrations to 32 consistently improves performance across all datasets. Error reflection continues to provide added benefit even at this larger scale. Notably, \method{} with error reflection and 8 demonstrations performs competitively with the 32-demo version without reflection, but at significantly lower cost. This highlights a potential cost-performance trade-off between scaling demonstrations and applying error reflection, which we leave for future work. 

Compared to five fine-tuning baselines, \method{} achieves comparable or superior performance on four out of five datasets, e.g., surpassing GoLLIE-34B on CoNLL03 and OntoNotes. These results suggest that the reliance on task-specific NER models may diminish, as general-purpose LLMs become increasingly cost-effective and reusable across tasks, while specialized models incur additional maintenance and deployment overheads.

\subsection{Ablation Studies \& Analyses}
\label{sec:ablation}



\paragraph{Label-guided Retriever}  
To better understand the effectiveness of our retriever, we compare three retrieval methods across all five datasets using \texttt{GPT-4o}: sentence embeddings from \kate{}, our proposed token-based retriever, and an unweighted variant that excludes label-guided token statistics. All embeddings are generated using OpenAI's \texttt{text-embedding-3-small} model. As shown in Table~\ref{tab:retriever}, our label-guided method consistently outperforms the other two. The unweighted variant performs similarly to or worse than sentence embeddings, underscoring the importance of incorporating label-grounded, token-level weighting in retrieval. Table~\ref{tab:retriever_analysis} further provides a sanity check on demonstration retrieval quality: for entities that appear in both the training and test sets, our retriever includes a substantially higher proportion of these entities within the retrieved demonstrations compared to task-agnostic sentence embeddings, validating our argument about the inefficiency of generic retrievers. Further analysis of the two components in our retriever is provided in Appendix~\ref{sec:ablation_appendix}.

\begin{table}[h!]
\small
\begin{center}
\scalebox{0.9}{
\begin{tabular}{lcccc}
\toprule
\textbf{Dataset} & Sentence & Token (uwg.) & Token (wg, ours)  \\
\midrule
NCBI  & 80.0 & 80.2 & \textbf{82.7} \\
bc2gm  & 73.6 & 71.6 & \textbf{74.3} \\
CoNLL03  & 93.1 & 92.6 & \textbf{93.5}  \\
OntoNotes  & 80.7 & 81.8 & \textbf{83.5} \\
TweetNER7  & 62.2 & 62.7 & \textbf{63.6}  \\
\bottomrule
\end{tabular}
}
\end{center}
\caption{\label{tab:retriever} 
Comparison of three demonstration retrievers: sentence embeddings (\kate{}), our label-guided token-based retriever, and an unweighted variant without label statistics. Embeddings are generated using OpenAI’s \texttt{text-embedding-3-small}. Our method consistently outperforms the others, highlighting the value of label-grounded, token-level weighting.}
\end{table}

\begin{table}[h!]
\small
\begin{center}
\scalebox{1.0}{
\begin{tabular}{lcc}
\toprule
\textbf{Dataset} & Sentence & Token (wg, ours) \\
\midrule
NCBI  & 85.0 & 93.3 \\
bc2gm  & 72.5 & 83.1  \\
CoNLL03  & 70.1 & 74.6 \\
OntoNotes  & 62.7 & 80.6 \\
TweetNER7  & 76.4 & 77.6 \\
\bottomrule
\end{tabular}
}
\end{center}
\caption{\label{tab:retriever_analysis} Sanity check for demonstration retrievers. For entities appearing in both the training and test sets, we compare different retrievers based on the proportion of these entities contained within the retrieved eight demonstrations. Our retriever yields substantially more relevant entities than task-agnostic sentence embedding baselines, supporting our argument regarding the inefficiency of generic retrievers.}
\end{table}

\begin{figure}[!t]
\begin{center}
  \includegraphics[width=0.50\textwidth]
  {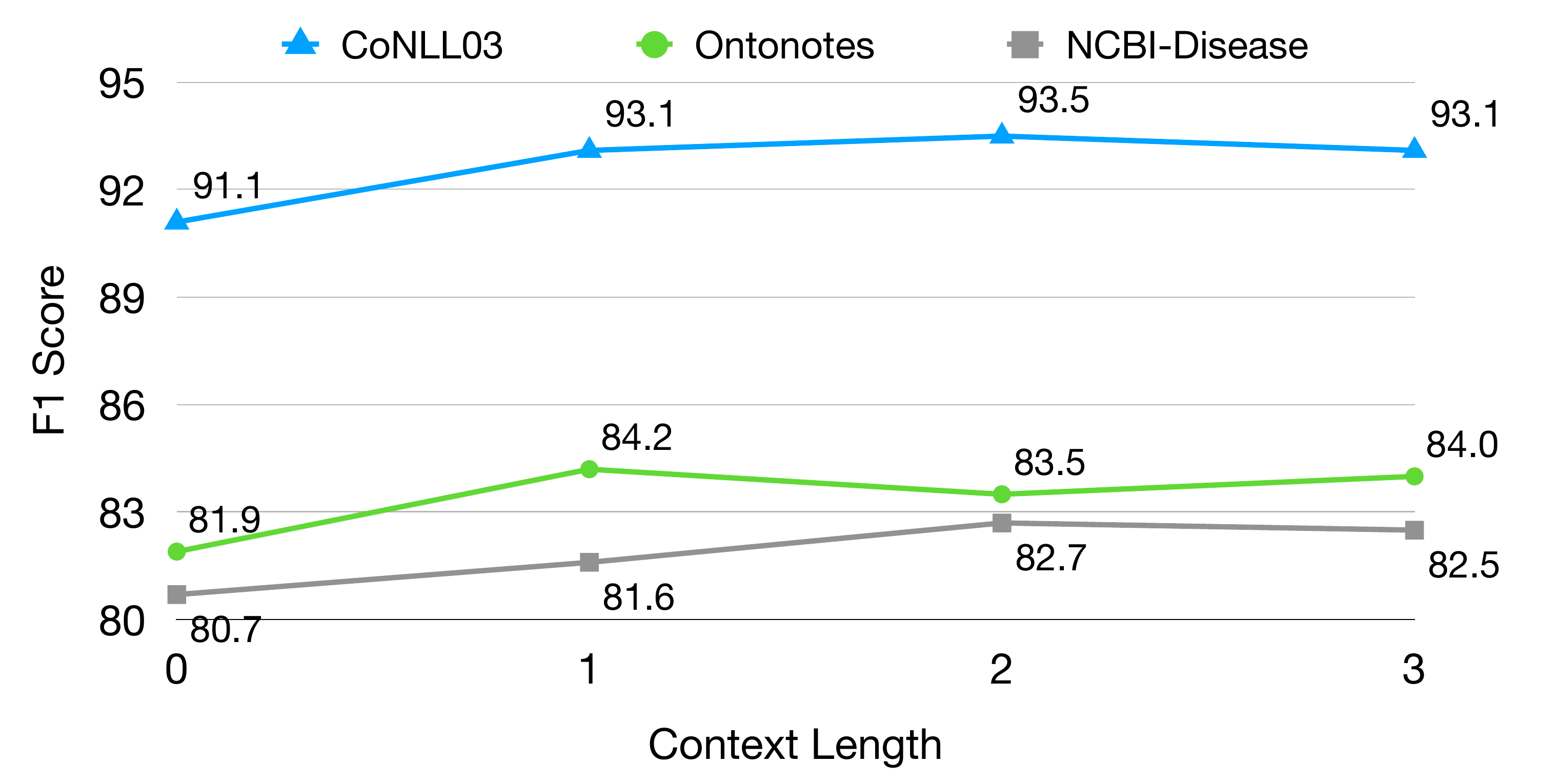}
\caption{Comparison of different context lengths in \method{} across three datasets, where 0 indicates that tokens are classified as either \textcolor{customlightblue}{entity} or \textcolor{green}{other} tokens. The substantial performance gain from context length 1 over 0 highlights the effectiveness of incorporating \textcolor{custompink}{context tokens} in our approach.
}
  \label{fig:context_len}
\end{center}
\end{figure}

\paragraph{Token Types \& Context Length}  
A key contribution of \method{} is the introduction of three token types, particularly \textcolor{custompink}{context tokens}, defined as tokens surrounding annotated entities within a specified context window. 
We conduct an ablation study on different context lengths, ranging from 0 to 3, where 0 indicates the removal of \textcolor{custompink}{context tokens}, leaving tokens classified only as \textcolor{customlightblue}{entity} or \textcolor{green}{other} tokens. The ICL results of \method{} on three datasets, presented in Figure~\ref{fig:context_len}, show that incorporating \textcolor{custompink}{context tokens} significantly boosts performance, although the improvement plateaus once the context length reaches 2. Moreover, Table~\ref{tab:unseen_appendix} in the appendix provides a breakdown of performance on seen and unseen entities as in Table~\ref{tab:unseen}, confirming that \textcolor{custompink}{context tokens} are effective in improving performance on unseen entities, supporting our hypothesis.

\paragraph{Low-Resource Results}  
Thus far, our evaluation has focused on high-resource settings, where training sets contain at least 5,000 examples. To assess \method{}'s effectiveness in low-resource scenarios, we downsample NCBI to three smaller scales: 16, 64, and 256 examples, and compare its performance against \kate{}. The results in Table~\ref{tab:low_resource} show that \method{} consistently outperforms \kate{}, demonstrating its robustness and effectiveness in low-resource conditions.

\begin{table}[h!]
\small
\begin{center}
\scalebox{1}{
\begin{tabular}{lcccc}
\toprule
\textbf{Method} & 16-shot & 64-shot & 256-shot  \\
\midrule
\kate{} & 70.3 & 73.2 & 73.6 \\
\method{} & \textbf{75.7} & \textbf{76.0} & \textbf{78.1} \\
\bottomrule
\end{tabular}
}
\end{center}
\caption{\label{tab:low_resource} Low-resource experiments on NCBI. We compare \method{} with \kate{} (using \texttt{GPT-4o}) across three scales: 16, 64, and 256 examples. \method{} surpasses \kate{} in all settings, highlighting its effectiveness in low-resource settings.}
\end{table}

\section{Related Work}
\label{sec:related_work}

\paragraph{LLM-Based Named Entity Recognition}  
The rise of LLMs has opened new avenues for NER by leveraging their capabilities in instruction-following \citep{ouyang2022traininglanguagemodelsfollow} and in-context learning \citep{brown2020languagemodelsfewshotlearners}. 
Recent work on LLM-based NER follows two main directions. One focuses on training NER-specialized models by fine-tuning open-source LLMs \citep{wang2023instructuie, zhou2024universalner, sainz2024gollieannotationguidelinesimprove}. The other, which we follow, explores adapting general-purpose LLMs for NER via prompting \citep{xie-etal-2023-empirical, ashok2023promptnerpromptingnamedentity, han2024empiricalstudyinformationextraction}.
In this direction, the dominant framework, \kate{} \citep{liu-etal-2022-makes}, selects demonstrations using sentence embeddings. Prior research mainly varies prompting templates \citep{li-etal-2023-codeie} or embedding models for retrieval \citep{monajatipoor2024llmsbiomedicinestudyclinical}, without addressing the core limitations of \kate{} in leveraging labeled data for token-level tasks. We address this gap by introducing a label-guided retriever and an error reflection strategy. The most relevant prior work is \citet{wang2023gptnernamedentityrecognition}, which proposes an entity-based retriever that requires training a BERT-based NER model to guide ICL retrieval. However, the resulting ICL model underperforms the BERT-based retriever itself, limiting its practical value. In contrast, our method shows that general-purpose LLMs can achieve competitive performance with supervised models in a training-free manner.

\paragraph{Reflection Using LLMs}
An emerging capability of LLMs is their ability to reason \citep{wei2023chainofthoughtpromptingelicitsreasoning} and self-reflect \citep{asai2023self} on their predictions, enabling them to correct errors. Recent research has explored this capability across a wide range of tasks, like code generation \citep{madaan2023selfrefineiterativerefinementselffeedback} and planning \citep{shinn2023reflexionlanguageagentsverbal}. Similar approaches have also been applied to NER \citep{wang2023gptnernamedentityrecognition, chen-etal-2024-double}. These works primarily rely on LLMs' internal knowledge for decision-making. In contrast, our approach grounds model predictions to the label-grounded training statistics. This grounding enables the model to capture fine-grained dataset-specific artifacts, akin to supervised methods, and helps close the performance gap between prompting-based and supervised approaches.

\section{Conclusion}


We present \textbf{\textsc{\method{}}}, a training-free ICL method designed to help LLMs make more informed entity predictions. \method{} leverages label statistics to identify tokens that are most informative for entity recognition, thereby improving demonstration retrieval and enabling effective error reflection. Evaluated on five NER datasets across four LLMs, \method{} consistently outperforms existing ICL approaches and achieves performance comparable to supervised fine-tuning. Further analyses highlight the limitations of commonly used sentence embedding-based retrievers and demonstrate \method{}’s effectiveness on both seen and unseen entities, as well as its robustness in low-resource settings.

\section*{Limitations}
\label{sec:limitations}

This paper focuses on flat NER while acknowledging that other NER challenges, such as discontinuous NER \citep{dai-etal-2020-effective} and nested NER \citep{finkel2009nested}, pose unique difficulties, particularly for autoregressive LLMs. Exploring these more complex NER tasks is left for future work.  
Our method primarily targets demonstration selection within the in-context learning framework. Future research could extend this by jointly optimizing both demonstration selection and task instructions \citep{pang-etal-2023-guideline}, as incorporating richer task-specific prompts may improve generalization in more difficult scenarios.
Finally, the error reflection mechanism used in this work is guided by domain knowledge and limited to three predefined error types. Recent studies have shown that LLMs are capable of detecting their own errors \citep{wang2024promptagent, kamoi2024evaluatingllmsdetectingerrors}, suggesting the potential for automating error reflection using more fine-grained or adaptive error categories. 



\bibliography{custom}

\appendix

\clearpage

\section{Implemetation Details}
\label{sec:implement_details}

\subsection{\method{} Prompts}
\label{sec:full_prompt}

We use the CoNLL03 dataset \citep{tjong-kim-sang-de-meulder-2003-introduction}, which targets the identification of "Person," "Organization," "Location," and "Miscellaneous" entities, as an example to illustrate the prompts used in our experiments.

\textbf{In-Context Learning - Input}

\begin{lstlisting}
Here is the JSON template for named entity recognition:
{"named entities": [{"name": "ent_name_1", "type": "ent_type_1"}, ..., {"name": "ent_name_n", "type": "ent_type_n"}]}

Please identify the four types of named entities: "PER", "LOC", "ORG", and "MISC" ("Miscellaneous"), following the JSON template listed above, and output the JSON object. If no named entities identified, output {"named entities": []}.

Input: The girl , who was accompanied to Philadelphia by her parents , will need more surgery later to correct the condition on her chest , back and legs , the hospital said .
Output: {"named entities": [{"name": "Philadelphia", "type": "LOC"}]}

Input: " I know what I 'm here for , " said Medvedev , who lost in the second round of the Open the last two years after reaching the quarters in 1993 , the same year he tried his hand as a restaurant critic .
Output: {"named entities": [{"name": "Medvedev", "type": "PER"}, {"name": "Open", "type": "MISC"}]}

Input: The church in Australia said on Monday Lynch , Batchelor , Barton and Riel were held in a prison until the weekend , when they were moved to join the other captives at the compound .
Output: {"named entities": [{"name": "Australia", "type": "LOC"}, {"name": "Lynch", "type": "PER"}, {"name": "Batchelor", "type": "PER"}, {"name": "Barton", "type": "PER"}, {"name": "Riel", "type": "LOC"}]}

Input: In a telephone call to a local newspaper from his holiday home in Spain , Dalglish said : " We came to the same opinion , albeit the club came to it a little bit earlier than me . "
Output: {"named entities": [{"name": "Spain", "type": "LOC"}, {"name": "Dalglish", "type": "PER"}]}

Input: Bosnian refugees in Hungary , the first to vote last weekend in their country 's first post-war election , found the rules confusing and some had no idea who they voted for , refugees and officials said on Wednesday .
Output: {"named entities": [{"name": "Bosnian", "type": "MISC"}, {"name": "Hungary", "type": "LOC"}]}

Input: Glasgow Rangers striker Ally McCoist , another man in form after two hat-tricks in four days , was also named for the August 31 World Cup qualifier against Austria in Vienna .
Output: {"named entities": [{"name": "Glasgow Rangers", "type": "ORG"}, {"name": "Ally McCoist", "type": "PER"}, {"name": "World Cup", "type": "MISC"}, {"name": "Austria", "type": "LOC"}, {"name": "Vienna", "type": "LOC"}]}

Input: Austrian television said the coach , which was carrying 45 , was en route from the Czech Republic to Italy when the accident occurred near Steinberg , 200 km southwest of Vienna .
Output: {"named entities": [{"name": "Austrian", "type": "MISC"}, {"name": "Czech Republic", "type": "LOC"}, {"name": "Italy", "type": "LOC"}, {"name": "Steinberg", "type": "LOC"}, {"name": "Vienna", "type": "LOC"}]}

Input: Austrian television reported earlier that more than 20 had been hurt in the accident at the station in Linz , 300 km ( 180 miles ) west of Vienna .
Output: {"named entities": [{"name": "Austrian", "type": "MISC"}, {"name": "Linz", "type": "LOC"}, {"name": "Vienna", "type": "LOC"}]}

Input: The fans , in Austria to watch their team play Rapid Vienna last Wednesday , may have been involved in a pub brawl earlier , the spokeswoman said .
Output: 
\end{lstlisting}

\textbf{In-Context Learning - Output}

\begin{lstlisting}
{"named entities": [{"name": "Austria", "type": "LOC"}, {"name": "Rapid Vienna", "type": "ORG"}]}
\end{lstlisting}

\textbf{Reflect Unseen - Input}

\begin{lstlisting}
<input_text>
Bitar pulled off fine saves whenever they did .
</input_text>

<candidate_tokens>
['Bitar']
</candidate_tokens>

<candidate_token>
Bitar
</candidate_token>
<potential_context_tokens_around>
['pulled']
</potential_context_tokens_around>
<context_token>
pulled
</context_token>
<examples>
Input: 10th-ranked American Chanda Rubin has pulled
Output: {"name": "Chanda Rubin", "type": "PER"}
</examples>

Given the input text from a Reuters news article, evaluate each candidate token along with the surrounding tokens to determine if it should be categorized as (part of) one of the four named entity types: "PER" (Person), "LOC" (Location), "ORG" (Organization), or "MISC" (Miscellaneous). **Use provided examples, if available, for reference.** If it should be a new entity, extract the exact text span in the sentence, including any spaces, in JSON format. Ensure tokens are not split (e.g., maintain hyphenated words intact). Note that abbreviations and full names are separate entities. If no changes are made, return {"named entities": []}.

JSON Template:
{"named entities": [{"name": "ent_name_1", "type": "ent_type_1"}, ..., {"name": "ent_name_n", "type": "ent_type_n"}]}

Output Format:

Candidate Token: token_1
Contextual Meaning: ...
Relation to Examples Provided: ...
Rationale: ...
Updates: ... (add a new entity/no change)

Candidate Token: token_n
Contextual Meaning: ...
Relation to Examples Provided: ...
Rationale: ...
Updates: ... (add a new entity/no change)

Final predicted entities for the input text (JSON format):
\end{lstlisting}

\textbf{Reflect Unseen - Output}

\begin{lstlisting}
Candidate Token: Bitar  
Contextual Meaning: "Bitar" appears to be a proper noun, likely referring to a person involved in a sporting context, as indicated by the phrase "pulled off fine saves."  
Relation to Examples Provided: The example provided features a person (Chanda Rubin) and categorizes her as "PER." This aligns with the context of "Bitar," suggesting it is also a person's name.  
Rationale: Given that "Bitar" is a proper noun and is likely a person's name in a sports context, it should be categorized as a person entity.  
Updates: Add a new entity.

Final predicted entities for the input text (JSON format):  
{"named entities": [{"name": "Bitar", "type": "PER"}]}
\end{lstlisting}

\textbf{Reflect False Negative - Input}

\begin{lstlisting}
<input_text>
Italian Serie A games to be played on Sunday ( league positions in parentheses , all kick- off times GMT ) :
</input_text>

<candidate_tokens>
['Italian']
</candidate_tokens>

<candidate_token>
Italian
</candidate_token>
<token_stat>
{"num_occurrences_as_entity": 35, "num_occurrences_as_context_tokens": 0, "num_occurrences_as_other_tokens": 0, "entity_vs_context_count": "35 vs 0", "entity_vs_non_entity_count": "35 vs 0"}
</token_stat>
<examples>
Positive Examples (part of entity):
Input: Middlesbrough 's Italian striker Fabrizio
Output: {"name": "Italian", "type": "MISC"}

Hard Negative Examples (context tokens):


Negative Examples (other tokens, not entity nor context):

</examples>

Please follow the instructions below:
1. Evaluate each candidate token listed above to determine if it should be categorized as (part of) one of the four named entity types: "PER" (Person), "LOC" (Location), "ORG" (Organization), or "MISC" (Miscellaneous). Consider both the positive and negative examples provided carefully. Pay particular attention to the overall statistical data on whether tokens are included or excluded from the entity. Hard negative examples highlight tokens that are not part of the entity but are located near it.
2. Review both sets carefully. In many cases, a token may be identified as part of an entity in positive examples but not in negative ones, likely due to inconsistencies in the annotation process. if positive and (hard) negative examples seem similar, **base your decision on statistical data**, such as the frequency of the span being recognized as an entity versus its context, particularly when the data is clear-cut (e.g., one frequency is significantly higher).
3. If the token has not been seen or is rarely seen in the training data, use your best judgment to determine whether it should be considered as part of or the entire name of a **specific** entity.
4. If any modifications are necessary, provide the updated entities by **extracting the exact text span in the sentence**, including any spaces and no outside tokens added, in JSON format. Ensure tokens are not split (e.g., maintain hyphenated words intact). Note that abbreviations and full names are separate entities. If no changes are required, return {"named entities": []}.

JSON Template:
{"named entities": [{"name": "ent_name_1", "type": "ent_type_1"}, ..., {"name": "ent_name_n", "type": "ent_type_n"}]}

Output Format:

Candidate Token: token_1
Training Data Stats: ... entity ... context ... regular ...
Contextual Meaning: ...
Relation to Examples Provided: ... positive examples ... negative examples ...
Rationale: ...
Updates: ... (add a new entity/no change)

Candidate Token: token_n
Training Data Stats: ... entity ... context ... regular ...
Contextual Meaning: ...
Relation to Examples Provided: ... positive examples ... negative examples ...
Rationale: ...
Updates: ... (add a new entity/no change)

Final predicted entities for the input text (JSON format):
\end{lstlisting}

\textbf{Reflect False Negative - Output}

\begin{lstlisting}
Candidate Token: Italian  
Training Data Stats: {"num_occurrences_as_entity": 35, "num_occurrences_as_context_tokens": 0, "num_occurrences_as_other_tokens": 0, "entity_vs_context_count": "35 vs 0", "entity_vs_non_entity_count": "35 vs 0"}  
Contextual Meaning: Refers to something related to Italy, in this case, the Italian Serie A football league.  
Relation to Examples Provided: Positive examples indicate that "Italian" is recognized as part of an entity (MISC) in contexts related to football or culture. There are no hard negative examples that contradict this.  
Rationale: The statistical data strongly supports that "Italian" is consistently recognized as an entity, specifically as a miscellaneous term related to the context of sports.  
Updates: No change  

Final predicted entities for the input text (JSON format):  
{"named entities": [{"name": "Italian", "type": "MISC"}]}
\end{lstlisting}

\textbf{Reflect Boundary - Input}

\begin{lstlisting}
<input_text>
Xinhua did not say when Qinglan port in Wenchang city would be opened to foreign vessels .
</input_text>

<predicted_entity>
{"name": "Wenchang city", "type": "LOC"}
</predicted_entity>

<boundary_tokens>
['city']
</boundary_tokens>

<boundary_token>
city
</boundary_token>
<status>
part of the entity
</status>
<token_stat>
{"num_occurrences_as_entity": 0, "num_occurrences_as_context_tokens": 44, "num_occurrences_as_other_tokens": 20, "entity_vs_context_count": "0 vs 44", "entity_vs_non_entity_count": "0 vs 64"}
</token_stat>
<examples>
Positive Examples (part of entity):


Hard Negative Examples (context tokens):
Input: in the English city of
Output: {"name": "English", "type": "MISC"}

Negative Examples (regular tokens, neither entity nor context):
Input: said the city council would
Output: {}
</examples>

Please follow the instructions below:
1. Calibrate the boundary of the predicted entity by evaluating each boundary token listed above in relation to the predicted entity. Don't consider it if it belongs to adjacent entities. Check both the provided positive and negative examples, with particular attention to the context surrounding the boundary token and overall statistical data on inclusion or exclusion from the entity. Hard negative examples highlight tokens that are not part of the entity but are located near it.
2. Review both sets carefully. In many cases, a token may be identified as part of an entity in positive examples but not in negative ones, likely due to inconsistencies in the annotation process. if positive and (hard) negative examples seem similar, **base your decision on statistical data**, such as the frequency of the span being recognized as an entity versus its context, particularly when the data is clear-cut (e.g., one frequency is significantly higher).
3. If the token has not been seen or is rarely seen in the training data, use your best judgment to determine whether it should be considered as part of or the entire name of a **specific** entity.
4. Determine whether any modifications, such as adding or removing boundary tokens, are necessary. If changes are required, provide the updated entity by **extracting the exact text span in the sentence**, including any spaces and no outside tokens added, in JSON format. Note that abbreviations and full names are separate entities. If no tokens are added to or removed from the predicted entity, output the original entity. If all original tokens are removed, output {}.

JSON Template:
{"name": "ent_name", "type": "ent_type"}

Output Format:

Boundary Token: token_1
Training Data Stats: ... entity ... context ... regular ...
Contextual Meaning: ...
Rationale: ... positive examples ... negative examples ... data stats ...

Boundary Token: token_n
Training Data Stats: ... entity ... context ... regular ...
Contextual Meaning: ...
Rationale: ... positive examples ... negative examples ... data stats ...

Updated Predicted Entity (JSON format):
\end{lstlisting}

\textbf{Reflect Boundary - Output}

\begin{lstlisting}
Boundary Token: city  
Training Data Stats: 0 entity, 44 context, 20 regular  
Contextual Meaning: The term "city" is used to describe a type of location but does not specify a unique entity on its own.  
Rationale: In the provided examples, "city" appears in contexts that do not contribute to identifying a specific entity. The positive examples do not include "city" as part of the entity, while the hard negative examples show that "city" is used in a general context. The statistical data indicates that "city" has not been recognized as part of an entity in the training data.  

Updated Predicted Entity (JSON format):  
{"name": "Wenchang", "type": "LOC"}
\end{lstlisting}

\begin{table}[h!]
\small
\begin{center}
\scalebox{0.9}{
\begin{tabular}{lccccc}
\toprule
 & \kate{} & \multicolumn{4}{c}{\textbf{\method{}}} \\
 \cmidrule(l){2-2}  \cmidrule(l){3-6} 
 & ICL & ICL & \multicolumn{3}{c}{Error Reflection} \\
  \cmidrule(l){2-2} \cmidrule(l){3-3} \cmidrule(l){4-6} 
\textbf{Dataset} & Emb. & Token & Unseen & FN & Bound. \\
\midrule
NCBI-disease  & 80.0 & 82.7 & 83.0 & 83.0 & 84.8 \\
bc2gm  & 73.6 & 74.3 & 74.3 & 74.4 & 75.4 \\
CoNLL03 & 92.9 & 93.5 & 93.5 & 93.5 & 93.6 \\
OntoNotes  & 80.7 & 83.5 & 83.4 & 83.6 & 85.2 \\
TweetNER7  & 62.2 & 63.6 & 63.8 & 63.8 & 64.3 \\
\bottomrule
\end{tabular}
}
\end{center}
\caption{\label{tab:breakdown} Performance breakdown of \method{} (with \texttt{GPT-4o}) by each step in its pipeline. The subsequent reflection steps demonstrate performance gains, validating the effectiveness of the proposed reflection strategy.}
\end{table}


\begin{table}[h!]
\small
\begin{center}
\scalebox{0.75}{
\begin{tabular}{ccccc}
\toprule
  & & & \multicolumn{2}{c}{\textbf{Unseen Ent.}} \\
   \cmidrule(l){4-5} 
 Dataset & Context & \multicolumn{1}{c}{\textbf{Seen Ent.}} & \multicolumn{1}{c}{Seen Tok.} & \multicolumn{1}{c}{Unseen Tok.} \\
\midrule
 \multirow{4}{*}{NCBI-disease} & 0 & 90.4 & 59.8 & 70.5 \\
 & 1 & 90.4 & 62.1 & 73.3 \\
 & 2 & 90.6 & 64.1 & \textbf{76.8} \\
 & 3 & \textbf{90.8} & \textbf{65.4} & 73.8  \\
\midrule
 \multirow{4}{*}{CoNLL03} & 0 & 95.5 & 84.9 & 91.2 \\
 & 1 & 95.6 & 84.8 & 91.7 \\
 & 2 & \textbf{95.7} & \textbf{86.9} & 92.0 \\
 & 3 & 95.2 & 86.2 & \textbf{92.1} \\
 
\midrule
 \multirow{4}{*}{OntoNotes} & 0 & 88.1 & 63.7 & 76.2\\
 & 1 & \textbf{89.5} & 67.5 & \textbf{80.2} \\
 & 2 & 88.6 & 67.8 & 79.6  \\
 & 3 & 89.4 & \textbf{68.3} & 78.9\\
\bottomrule
\end{tabular}
}
\end{center}
\caption{\label{tab:unseen_appendix} Performance breakdown of \method{} on seen and unseen entities with varying context lengths (where 0 indicates the removal of \textcolor{custompink}{context tokens}). The results demonstrate that \textcolor{custompink}{context tokens} significantly enhance \method{}'s performance, particularly on unseen entities.}
\end{table}

The hyper-parameters used for \method{} are presented in Table \ref{tab:hyper_deer}, and the hyper-parameters used for fine-tuning BERT-base are presented in Table \ref{tab:hyper_bert}.

\begin{table}[h!]
\begin{center}
\scalebox{0.7}{
\begin{tabular}{lccccccccc}
\toprule
 Dataset & \( \lambda_1 \) & \( \lambda_2 \) & $C$ & $w_{e}$ & $w_{c}$ & $w_{o}$ & $\theta_{FN}$ & $M$ & $K$ \\
 \midrule
NCBI & 1 & 1 & 2 & 1.0 & 1.0 & 0.01 & 0.95 & 1 & 2 \\
bc2gm & 1 & 0.01 & 2 & 1.0 & 0.5 & 0.01 & 0.9 & 1 & 2 \\
CoNLL03 & 0.01 & 1 & 2 & 1.0 & 1.0 & 0.01 & 0.95 & 1 & 2 \\
OntoNotes & 1 & 1 & 2 & 1.0 & 1.0 & 0.01 & 0.95 & 4 & 2 \\
TweetNER7 & 1 & 1 & 2 & 1.0 & 1.0 & 0.01 & 0.95 & 1 & 2 \\
\bottomrule
\end{tabular}
}
\end{center}
\caption{\label{tab:hyper_deer} Hyper-parameters for \method{}. \( \lambda_1 \) and \( \lambda_2 \) control the relative contribution of two components in the retrieval. $C$ refers to context length. $w_{e}$, $w_{c}$, and $w_{o}$ weight the relative importance of three token types. $\theta_{FN}$ is set as a threshold for reflections on false negative tokens. $M$ is the number of span-level demonstrations in the reflection prompt. $K$ is the number of tokens for boundary reflection.}
\end{table}

\begin{table}[h!]
\begin{center}
\scalebox{0.8}{
\begin{tabular}{cc}
\toprule
learning rate & $1\times10^{-5}$ \\
\# epoch & 10 \\
batch size & 16 \\
gradient accumulation steps & 2 \\
\bottomrule
\end{tabular}
}
\end{center}
\caption{\label{tab:hyper_bert} Hyper-parameters for BERT-base fine-tuning.}
\end{table}

\begin{table}[h!]
\small
\begin{center}
\scalebox{0.9}{
\begin{tabular}{lccccc}
\toprule
\textbf{Dataset} & Unseen & FN & Bound. \\
\midrule
NCBI  & 92 & 20 & 130 \\
bc2gm  & 23 & 4 & 163 \\
CoNLL03 & 38 & 12 & 29\\
OntoNotes  & 35 & 21 & 115 \\
TweetNER7  & 174 & 6 & 215 \\
\bottomrule
\end{tabular}
}
\end{center}
\caption{\label{tab:reflect_freq} The number of reflections triggered across five datasets using \texttt{GPT-4o}. The majority of reflections are concentrated on boundary tokens, while the number of reflections involving unseen tokens remains relatively small and computationally manageable.}
\end{table}

\subsection{Statistical Test on Sampled Datasets}
\label{sec:stat_test}

To ensure the sampled subset provides a reliable estimate of model performance, we calculate 95\% confidence intervals for \method{}'s F1 scores using 1,000 bootstrap resamples. The margins of error for the three datasets are shown in Table~\ref{tab:stat_test} and fall within a 0.01-0.03 range, suggesting that the random sampling does not significantly affect the validity of our performance estimates.

\begin{table}[h!]
\small
\begin{center}
\scalebox{1.0}{
\begin{tabular}{lccccc}
\toprule
\textbf{Dataset} & Error Margin \\
\midrule
bc2gm  & 0.027 \\
CoNLL03 & 0.013 \\
OntoNotes  & 0.019 \\
\bottomrule
\end{tabular}
}
\end{center}
\caption{\label{tab:stat_test} Margins of error for three sampled datasets.}
\end{table}

\begin{table}[h!]
\small
\begin{center}
\scalebox{0.8}{
\begin{tabular}{lccccc}
\toprule
\textbf{Method} & BERT\textsubscript{base}	& Qwen2.5\textsubscript{7B} & \texttt{GPT-4o-mini} & \texttt{GPT-4o} \\
\midrule
Fine-tune  & 84.0 & - & - & - \\
\kate{} & - & 66.9 & 76.0 & 80.0 \\
GPT-NER  & - & 71.2 & 75.6 & 82.7 \\
\method{} (w/o ER) & - & \textbf{71.7} & \textbf{77.6} & \textbf{82.7} \\
\bottomrule
\end{tabular}
}
\end{center}
\caption{\label{tab:gpt_ner} Comparing our \method{} with GPT-NER. \method{} consistently outperforms GPT-NER across different LLMs. Notably, GPT-NER even underperforms its original BERT-based retriever, highlighting its limitations and impracticality.}
\end{table}

\section{Ablation Studies}
\label{sec:ablation_appendix}

In addition to the ablation studies presented in \S\ref{sec:ablation}, we further investigate the impact of other components of \method{} below. 

\begin{table}[h!]
\small
\begin{center}
\scalebox{0.75}{
\begin{tabular}{lcccc}
\toprule
 &  & Tagging & Python func. & JSON  \\
\textbf{Dataset} & LLM &  \citep{paolini2021structuredpredictiontranslationaugmented} & \citep{li-etal-2023-codeie} & (ours)  \\
\midrule
\multirow{2}{*}{NCBI} & \texttt{GPT-4o-mini} & 71.3 & 74.8 & 77.6 \\
     & \texttt{GPT-4o} & 80.1 & 81.7 & 82.7 \\
\bottomrule
\end{tabular}
}
\end{center}
\caption{\label{tab:format} Comparing our JSON output format with prompting formats from recent works \citep{paolini2021structuredpredictiontranslationaugmented, li-etal-2023-codeie} for the ICL step in \method{}, using \texttt{GPT-4o-mini} and \texttt{GPT-4o}. Our JSON format consistently outperforms the other two prompting formats, though the performance gap narrows when using \texttt{GPT-4o}.}
\end{table}

\subsection{Prompting Template} 

We compare our JSON output format \citep{dunn2022structuredinformationextractioncomplex, lu-etal-2022-unified, bai-etal-2024-schema} with other prompting formats explored in previous studies, including tagging-based formats \citep{paolini2021structuredpredictiontranslationaugmented, wang2023gptnernamedentityrecognition}, such as \textit{``[Barack Obama | PERSON] was ...''}, and Python function generation \citep{li-etal-2023-codeie}. 
Results in Table~\ref{tab:format} show that our selected format outperforms two other formats, though the performance gap narrows when using \texttt{GPT-4o}.

\subsection{Retrieval Embeddings}  

Following previous work \citep{monajatipoor2024llmsbiomedicinestudyclinical}, we experiment with seven embedding models to identify the most suitable sentence embeddings for \kate{}. These models include six open-source models available on HuggingFace\footnote{\url{https://huggingface.co/models}} via the \texttt{sentence-transformer} library\footnote{\url{https://sbert.net/}}, along with OpenAI's proprietary model \texttt{text-embedding-3-small}. For open-source models, we include both general-domain models, such as \texttt{all-mpnet-base-v2}, and domain-specific models, such as \texttt{BioClinicalBERT}. Table~\ref{tab:embedding_model} presents the performance of these seven embedding models with \kate{} using \texttt{GPT-4o-mini}. The results show that \texttt{BioClinicalBERT} outperforms other open-source models, highlighting the benefits of in-domain pre-training. Among all models, \texttt{text-embedding-3-small} achieves the best performance, surpassing others by a clear margin. Consequently, we adopt \texttt{text-embedding-3-small} for all datasets in our experiments.

\begin{table}[h!]
\small
\begin{center}
\scalebox{1}{
\begin{tabular}{lc}
\toprule
 Emb. Model & \kate{}  \\
\midrule
\texttt{bert-base-cased} & 70.9 \\
 \texttt{bert-base-uncased} & 72.8 \\
 \texttt{all-MiniLM-L6-v2} & 73.2 \\
 \texttt{all-mpnet-base-v2} & 73.3 \\
 \texttt{BioBERT} & 72.2 \\
 \texttt{BioClinicalBERT} & 73.6 \\
 \texttt{text-embedding-3-small} & \textbf{76.0} \\
\bottomrule
\end{tabular}
}
\end{center}
\caption{\label{tab:embedding_model} Comparing seven embedding models to select the most suitable sentence embeddings for \kate{} using \texttt{GPT-4o-mini}. \texttt{text-embedding-3-small} surpasses other models by a clear margin, thus adopted for all our experiments.}
\end{table}

\subsection{Contextualized v.s. Uncontextualized Embeddings}

We compare two types of token embeddings for our token-focused retriever. Contextualized token embeddings are obtained by averaging sub-token embeddings from the last layer of the embedding model, capturing sentence-level contextual information. Uncontextualized embeddings encode each token in the vocabulary independently. We evaluate both types using seven embedding models.  Note that OpenAI's \texttt{text-embedding-3-small} does not provide sub-token embeddings from input text, so we use it only for uncontextualized token embeddings. The generated token embeddings are then weighted by training set token-level statistics to produce sentence-level embeddings for demonstration retrieval, as introduced in \S\ref{sec:method}. The results show no consistent pattern among open-source models regarding which type of token embeddings performs best; the outcome varies depending on the model. However, among all models, the uncontextualized embeddings from \texttt{text-embedding-3-small} achieve the best performance.

\begin{table}[h!]
\small
\begin{center}
\scalebox{1}{
\begin{tabular}{lcc}
\toprule
 Emb. Model & Contex. & Uncontex. \\
\midrule
\texttt{bert-base-cased} & 74.9 & 74.9 \\
 \texttt{bert-base-uncased} & 73.9 & 75.9 \\
 \texttt{all-MiniLM-L6-v2} & 74.5 & 75.1\\
 \texttt{all-mpnet-base-v2} & 73.4 & 74.7 \\
 \texttt{BioBERT} & 74.1 & 74.6\\
 \texttt{BioClinicalBERT} & 75.9 & 74.7\\
 \texttt{text-embedding-3-small} & - & \textbf{76.3} \\
\bottomrule
\end{tabular}
}
\end{center}
\caption{\label{tab:token_embedding} Comparing contextualized and uncontextualized token embeddings used in \method{} on NCBI. No consistent pattern among open-source models is found regarding which type of token embeddings performs best. Uncontextualized embeddings from \texttt{text-embedding-3-small} achieve the best performance.}
\end{table}

\subsection{Two Components in Label-guided Retriever}  
We compare four retrieval methods: sentence embeddings used in \kate{} (\texttt{S-Emb}), our token-based retriever, and its two components, weighted token matching (\texttt{T-Match}) and weighted token embeddings (\texttt{T-Emb}). The results across three datasets are presented in Table~\ref{tab:retriever_component}. The results show that \texttt{T-Emb} consistently outperforms \texttt{S-Emb}, underscoring the importance of prioritizing entity- and context-related tokens in demonstration retrieval for NER. Combining \texttt{T-Emb} with \texttt{T-Match} improves performance, though \texttt{T-Match} alone lags behind \texttt{S-Emb} on CoNLL03. Further analysis reveals that this issue stems from short, non-language data in the dataset, such as news bylines (e.g., ``OMAHA 1996-12-06'') or sports score tables, as noted by \citet{liu-ritter-2023-conll}. These examples contain a high proportion of unseen tokens, reducing the effectiveness of \texttt{T-Match}, which relies on exact token matching. 

\begin{table}[h!]
\small
\begin{center}
\scalebox{0.95}{
\begin{tabular}{lcccc}
\toprule
 & \kate{} & \multicolumn{3}{c}{\method{}}  \\
 \cmidrule(l){2-2} \cmidrule(l){3-5} 
\textbf{Retriever} & S-Emb & T-Match & T-Emb & Combined  \\
\midrule
NCBI  & 80.0 & 82.0 & 81.4 & \textbf{82.7} \\
CoNLL03  & 92.9 & 92.2 & 93.3 & \textbf{93.5} \\
OntoNotes  & 80.7 & 82.6 & 82.3 & \textbf{83.5}  \\
\bottomrule
\end{tabular}
}
\end{center}
\caption{\label{tab:retriever_component} Comparison of four demonstration retrievers: sentence embeddings (\texttt{S-Emb}), two components of \method{} (\texttt{T-Emb} and \texttt{T-Match}), and their combination. \texttt{T-Emb} consistently outperforms \texttt{S-Emb}, highlighting the importance of prioritizing entity- and context-related tokens in retrieval. Combining \texttt{T-Emb} with \texttt{T-Match} further enhances performance.}
\end{table}

\subsection{Token Type Weights}  

As explained in \S\ref{sec:method}, we use hyperparameters $w_{e}$, $w_{c}$, and $w_{o}$ to control the relative importance of the three token types. Their values are determined through a grid search over three predefined values: $[1.0, 0.5, 0.01]$. Due to budget constraints, we select \( w_{e} = 1.0 \), \( w_{c} = 1.0 \), and \( w_{o} = 0.01 \) based on initial trials. This choice aligns with the intuition that entity- and context-related tokens are more important for NER. We find that these values perform well across most datasets. To evaluate the importance of each token type, we examine three sets of values, assigning $0.01$ to each weight in turn while keeping the others at $1.0$. Table~\ref{tab:token_weight} presents the results of the ICL step in \method{} under these three settings. The results show that the default set \( w_{e} = 1.0 \), \( w_{c} = 1.0 \), and \( w_{o} = 0.01 \) performs best, while the other two sets exhibit significant performance drops. This outcome verifies the critical importance of entity- and context-related tokens for NER performance.

\begin{table}[h!]
\small
\begin{center}
\scalebox{1}{
\begin{tabular}{lcccc}
\toprule
$w_{e}$ & $w_{c}$  & $w_{o}$  & \method{} \\
\midrule
0.01 & 1.0 & 1.0 & 71.6 \\
1.0 & 0.01 & 1.0 & 74.7 \\
1.0 & 1.0 & 0.01 & \textbf{77.6} \\
\bottomrule
\end{tabular}
}
\end{center}
\caption{\label{tab:token_weight} Comparing three sets of values, assigning $0.01$ to each weight in turn while keeping the others at $1.0$, to evaluate the importance of each token type. The default set \( w_{e} = 1.0 \), \( w_{c} = 1.0 \), and \( w_{o} = 0.01 \) performs best, verifying the importance of entity- and context-related tokens for NER performance.}

\end{table}

\begin{figure*}[ht!]
    \centering
    \includegraphics[width=\textwidth]{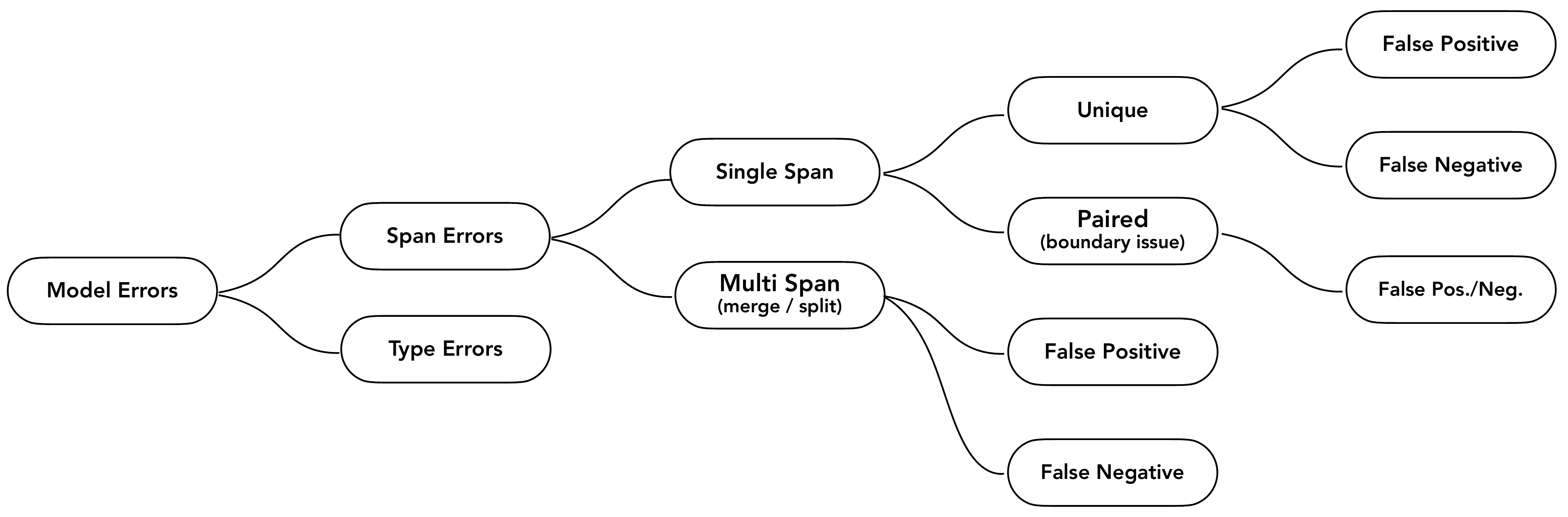}
    \caption{NER error ontology. This ontology is developed based on the standard strict mention-level matching metric for NER \citep{segura-bedmar-etal-2013-semeval}, where a predicted entity is considered correct only if both the predicted span boundaries and the entity type match with the gold ones. ``Span Errors'' refer to predictions that include at least span errors, while ``Type Errors'' denote cases where only the entity type is incorrect. ``Multi-Span Errors'' describe cases where multiple span issues occur, such as when a single gold entity is split into two predicted entities or when two gold entities are merged into one predicted entity.  Similar analyses can be found in recent works \citep{ding-etal-2024-rethinking, lu2024largelanguagemodelsstruggle}.
    }
    \label{fig:error_onto}
\end{figure*}

\begin{table*}[!ht]
\small
\centering
\resizebox{\linewidth}{!}{
\begin{tabular}{lcccccccccccccccccc}
\toprule
    & \multicolumn{6}{c}{\textbf{\kate{} (\texttt{GPT-4o})}} & \multicolumn{6}{c}{\textbf{\method{} (\texttt{GPT-4o})}} & \multicolumn{6}{c}{\textbf{BERT-base}} \\
 \cmidrule(l){2-7}  \cmidrule(l){8-13} \cmidrule(l){14-19}
 & Type & \multicolumn{5}{c}{Span} & Type & \multicolumn{5}{c}{Span} & Type & \multicolumn{5}{c}{Span} \\
 \cmidrule(l){2-2}  \cmidrule(l){3-7} \cmidrule(l){8-8}  \cmidrule(l){9-13} \cmidrule(l){14-14} \cmidrule(l){15-19}
 & & \multicolumn{2}{c}{Multiple} & \multicolumn{3}{c}{Single} & & \multicolumn{2}{c}{Multiple} & \multicolumn{3}{c}{Single} & & \multicolumn{2}{c}{Multiple} & \multicolumn{3}{c}{Single} \\
  \cmidrule(l){3-4}  \cmidrule(l){5-7} \cmidrule(l){9-10}  \cmidrule(l){11-13} \cmidrule(l){15-16} \cmidrule(l){17-19}
 & & & & \multicolumn{2}{c}{Unique} & Paired & &  & & \multicolumn{2}{c}{Unique} & Paired & & & & \multicolumn{2}{c}{Unique} & Paired \\
 \cmidrule(l){5-6} \cmidrule(l){7-7} \cmidrule(l){11-12} \cmidrule(l){13-13} \cmidrule(l){17-18} \cmidrule(l){19-19}
\textbf{Dataset} & \texttt{FP/FN} & \texttt{FP} & \texttt{FN} & \texttt{FP} & \texttt{FN} & \texttt{FP/FN} & \texttt{FP/FN} & \texttt{FP} & \texttt{FN} & \texttt{FP} & \texttt{FN} & \texttt{FP/FN} & \texttt{FP/FN} & \texttt{FP} & \texttt{FN} & \texttt{FP} & \texttt{FN} & \texttt{FP/FN} \\

\midrule

NCBI \citeyearpar{dougan2014ncbi} & 0 & 15 & 8 & 57 & 121 & 87 & 0 & 14 & 8 & 90 & 39 & 73 & 0 & 18 & 11 & 90 & 53 & 72 \\
bc2gm \citeyearpar{smith2008overview} & 0 & 48 & 77 & 90 & 115 & 160 & 0 & 59 & 89 & 154 & 48 & 130 & 0 & 98 & 75 & 103 & 63 & 97 \\
CONLL03 \citeyearpar{tjong-kim-sang-de-meulder-2003-introduction} & 53 & 4 & 8 & 31 & 44 & 21 & 54 & 3 & 6 & 40 & 23 & 23 & 66 & 17 & 16 & 50 & 28 & 38 \\
OntoNotes \citeyearpar{weischedel2013ontonotes} & 33 & 27 & 30 & 46 & 73 & 130 & 30 & 31 & 44 & 54 & 27 & 158 & 27 & 43 & 30 & 55 & 44 & 61 \\
TweetNER7 \citeyearpar{ushio-etal-2022-named} & 210 & 64 & 44 & 191 & 394 & 123 & 206 & 66 & 50 & 195 & 364 & 120 & 206 & 136 & 71 & 277 & 355 & 135 \\

\bottomrule
\end{tabular}
}

\caption{\label{tab:error_breakdown_full} Error breakdown for \kate{}, \method{}, and BERT-base based on the hierarchical ontology of NER errors (see Figure \ref{fig:error_onto}). The primary source of errors across all five datasets is span-related issues for single entities. \method{}, as intended by its design, performs better in reducing unique false negatives and paired errors compared to \kate{}. On the two datasets where \method{} lags behind BERT-base, the latter still exhibits superior boundary handling, suggesting the need for a better approach to further improve span prediction.
}
\end{table*}

\subsection{Llama3.1-8B Inadequacy}
\label{sec:llama_issue}
In early experiments with \texttt{Llama3.1-8B}, we observed frequent failures to generate outputs in the required JSON format during ICL inference. Table~\ref{tab:llama_issue} reports the percentage of test examples with format errors across five datasets. The failure rates range from 26.8\% to 82.7\%, making the model's outputs unreliable and difficult to evaluate. Due to these inconsistencies, we exclude \texttt{Llama3.1-8B} from our main experiments.

\begin{table}[h!]
\small
\begin{center}
\scalebox{1.0}{
\begin{tabular}{lccccc}
\toprule
\textbf{Dataset} & Failure Rate (\%) \\
\midrule
NCBI & 30.7 \\
bc2gm  & 26.8 \\
CoNLL03 & 81.3 \\
OntoNotes  & 82.7 \\
TweetNER7 & 67.5 \\
\bottomrule
\end{tabular}
}
\end{center}
\caption{\label{tab:llama_issue} Percentage of example outputs from \texttt{Llama3.1-8B} that failed to follow the required JSON format specified in the ICL prompt.}
\end{table}

\section{Error Analyses}
\label{sec:error}

We conduct an error breakdown based on the hierarchical ontology of NER errors in Figure \ref{fig:error_onto} for \kate{}, \method{}, and BERT-base. The full results are presented in Table \ref{tab:error_breakdown_full}, revealing that the primary source of errors across all five datasets is span-related issues for single entities, as highlighted in \S\ref{sec:method} when motivating the error reflection process. These errors consist of unique false positives and negatives, as well as paired errors, typically caused by boundary mismatches. \method{} outperforms \kate{} in reducing unique false negatives and paired errors, as intended by its design. 

\end{document}